\documentclass[opre,copyedit]{informs1_ajm}
\usepackage{lscape}
\usepackage{url}
\usepackage{xcolor}
\usepackage{graphicx}
\usepackage[mathscr]{euscript}
\usepackage[colorlinks=true, linkcolor=black, urlcolor=black, citecolor=black]{hyperref}
\usepackage{multirow}
\usepackage{array}
\usepackage{booktabs}
\usepackage{float}
\usepackage{psfrag}
\usepackage{bm}
\usepackage{tikz,pgf}
\usepackage{amsmath}
\usepackage{epsfig,fancybox}
\usepackage{amssymb}
\usepackage{etoolbox}
\usepackage[T1]{fontenc}
\usepackage{fix-cm}

\theoremstyle{EX}

\usepackage{endnotes}

\newcommand{\zee}{\hbox{{\sf Z}\kern-.3550em\hbox{{\sf Z}} }}
\newcommand{\real}{\hbox{{I\kern-.1667em\hbox{R}}}}

\newcommand{\ignore}[1]{}

\usepackage{natbib}
 \bibpunct[, ]{(}{)}{,}{a}{}{,}%
 \def\bibfont{\footnotesize}%
\TheoremsNumberedThrough     

\EquationsNumberedThrough    

\MANUSCRIPTNO{} 

\begin{document}

\TITLE{Tabular Foundation Models for Discrete Choice Estimation}

\ARTICLEAUTHORS{
\AUTHOR{Liu Liu, Dan Zhang}
\AFF{Leeds School of Business, University of Colorado Boulder \\
\EMAIL{liu.liu-1@colorado.edu, dan.zhang@colorado.edu} \URL{}}

}
\HISTORY{\today}

\ABSTRACT{Tabular foundation models (TFMs) generate predictions on structured data via
in-context learning, without task-specific estimation. We ask whether TFMs can be
effectively applied to discrete choice, a central demand estimation framework in marketing
and operations, and find that directly applying TFMs yields limited performance. The gap
is structural: TFMs assume row-independent observations, whereas
discrete choice is inherently set-valued and subject to persistent consumer preference
heterogeneity. We propose a reformulation that encodes both choice-set dependence and
individual heterogeneity within a row-based learning framework. Evaluated on a yogurt
scanner panel, individual-level heterogeneity encoding is the dominant driver of predictive
accuracy. The best reformulation outperforms hierarchical Bayesian estimation on both
holdout log-likelihood and hit rate, running 16 times faster, a practical
advantage for large-scale demand estimation. The advantage is largest in the medium-data
regime (10--40 purchase occasions per consumer), where parametric Bayesian shrinkage most
distorts estimates for atypical consumers. Fine-tuning on population choice data provides
additional gains for consumers with shallow purchase histories, where in-context learning
has limited individual-specific signal to condition on. These results establish a principled
approach for applying foundation models to consumer choice problems more broadly.
}

\maketitle

\section{Introduction}
\label{sec:intro}

Tabular foundation models (TFMs) have recently emerged as powerful pretrained architectures
for structured, column-based data \citep{hollmann2025accurate}. Like large language models
(LLMs) in text domains, TFMs are trained once on large collections of synthetic prediction
tasks and subsequently applied to new datasets via \emph{in-context learning}: just as an
LLM can perform a new task by conditioning on instructions or examples in the prompt at inference time rather than
retraining, a TFM produces predictions on a new dataset by treating the labeled training
examples as input context, with no task-specific parameter estimation or gradient updates
of any kind. Because inference consists of a single forward pass through a fixed pretrained
network, TFMs are computationally lightweight, and empirical evidence suggests they achieve
competitive predictive accuracy across a wide range of tabular benchmarks with minimal
tuning. These properties make them an appealing tool for empirical applications in marketing
and operations that involve structured, column-formatted data.

Discrete choice models (DCMs) constitute a central empirical framework in marketing,
economics, and operations management for analyzing demand and consumer decision-making
among competing alternatives. Under the random utility paradigm, each alternative in a
choice set is associated with a latent utility, and observed choices arise from utility
comparisons within the set. Choice probabilities are therefore defined conditional on the
composition of the choice set. Canonical models, including multinomial logit, mixed logit,
and latent class specifications, require explicit functional assumptions and are typically
estimated via likelihood-based procedures that may involve simulation and numerical
optimization \citep{train2009discrete}.

A central challenge in discrete choice estimation is preference heterogeneity: consumers
differ systematically in their tastes, and individual-level preference parameters are
typically not identified from the small number of observations available per person: often
ten to twenty tasks in conjoint studies or a few dozen purchase occasions in scanner panel
data. The hierarchical Bayesian (HB) approach has become the dominant solution in marketing
research and commercial practice \citep{rossi2005bayesian}. HB models a two-level
structure: individual preference parameters $\boldsymbol{\beta}_i$ are assumed drawn from a population
distribution, and observed choices are generated by utility maximization at the individual
level. Posterior distributions over individual parameters are recovered via Markov chain
Monte Carlo (MCMC), yielding estimates that are shrunk toward the population mean in
proportion to the information available for each consumer. The result is individual-level
preference estimates that support segmentation, targeting, and downstream optimization.

These considerations raise a natural question: can TFMs be effectively applied to discrete
choice problems, and can they match the predictive performance of HB without any
dataset-specific estimation? The answer is not obvious. Two structural features of choice
data sit in tension with the row-wise inductive bias of tabular learning, and reconciling
them is the central challenge we address.

A first source of tension is the set-valued nature of discrete choice. TFMs are pretrained
under a row-wise supervised learning paradigm in which each observation is treated as an
independent input-output pair. In contrast, discrete choice is inherently set-valued: the
probability of choosing an alternative is defined conditional on the composition of the
choice set and depends on comparisons across alternatives within that set. The alternatives
in a given choice task therefore form a mutually dependent group, linked through
substitution patterns and competitive interactions. Moreover, the identity of the chosen
alternative is invariant to the ordering of alternatives in the set, reflecting a
permutation invariance at the choice set level that is not imposed by standard tabular representations. Discrete
choice prediction therefore cannot be expressed as a function of an isolated row without
explicitly accounting for the relational structure of the choice set.

A second source of tension is preference heterogeneity across consumers. In most
applications, multiple choice tasks are observed for each decision maker, and these choices
are generated from a common underlying utility function. Observations associated with the
same individual are therefore not independent but reflect persistent heterogeneity in tastes
and substitution patterns. This feature contrasts with the pooled prediction setting
implicit in row-wise tabular learning, where all observations are treated as arising from a
single homogeneous mapping from features to outcomes. Taken together, discrete choice data
are inherently relational and conditional, violating row independence both within choice
sets and across observations belonging to the same individual.

We find that directly applying TFMs to discrete choice data yields limited predictive
performance, not from insufficient model capacity, but from this structural mismatch. To
address it, we propose a structured reformulation of discrete choice prediction for TFMs,
in which choice data are represented as tabular prediction tasks that explicitly encode
choice-set dependence and individual heterogeneity. We show that when these structural
properties are appropriately encoded, pretrained TFMs applied purely via in-context
learning match or exceed the predictive performance of HB while being substantially faster
to deploy.

The reformulation has two main components. First, a choice-set-to-tabular representation
that converts set-valued competition into row-level supervised inputs through set-aware and
pairwise constructions. Second, an individual heterogeneity encoding that captures consumer-specific preferences
either through an explicit respondent identifier feature or through per-respondent task
construction for in-context learning. We further consider permutation-based augmentation to approximate invariance with respect to alternative ordering.

We evaluate these representations on a yogurt scanner panel, a revealed-preference dataset
in which consumers made repeated brand choices under real market conditions. As the
representative TFM, we use TabPFN \citep{hollmann2023tabpfn, hollmann2025accurate}, a
pretrained transformer that achieves state-of-the-art accuracy on small-to-medium tabular
classification tasks while completing inference in seconds. The analysis
compares TabPFN against pooled MNL and hierarchical Bayesian benchmarks across a range of
consumer purchase histories, isolating the contributions of choice-set representation,
consumer heterogeneity encoding, and choice data adaptation to predictive
performance. We also evaluate fine-tuning as a route from the TFM's learned prior toward a
population choice-aware prior, following the framework developed in
Section~\ref{sec:tfm-shrinkage}.

This paper makes three contributions. First, we identify a structural mismatch between
discrete choice problems and TFMs, arising from the relational and conditional nature of
choice data and the row-wise inductive bias of tabular learning; we further develop a
unifying shrinkage perspective that connects HB and TFM in-context learning as instances
of the same adaptive pooling principle, differing only in how the preference prior is
acquired. Second, we propose a structured reformulation of discrete choice prediction for
TFMs, showing how appropriate data representation and task formulation can encode
choice-set structure and consumer heterogeneity within a row-wise learning framework.
Third, we provide a systematic empirical evaluation establishing three findings:
individual-level heterogeneity, not choice-set representation, is the dominant source of
predictive gain; appropriately reformulated TFMs match or exceed HB in predictive accuracy
at substantially lower computational cost, with the relative advantage greatest in the
medium-data regime; and fine-tuning on population choice data provides additional gains for
consumers with sparse purchase histories, where in-context learning has limited
individual-specific signal to condition on. More broadly, TFMs introduce a new source of prior for individual-level preference prediction from sparse data: one learned outside the focal sample, rather than estimated from it. Managerially, this makes TFMs an effective tool in precisely the settings that dominate practice, where each consumer supplies only a handful of purchase occasions.

Our analysis does not position TFMs as
substitutes for structural DCMs; classical approaches remain essential for interpretation
and counterfactual analysis. Nor does it propose a task-specific architecture in the spirit
of ML-for-choice methods, which require estimating parameters on the focal dataset.
Rather, we find that a fixed pretrained model, when the representation pipeline reflects
the structural features of choice data, serves as an effective and lightweight predictive
alternative to HB.

The remainder of the paper is organized as follows.
Section~\ref{sec:lit} reviews the related literature.
Section~\ref{sec:tfm-shrinkage} introduces the prior-data fitted network (PFN) framework
underlying TabPFN, develops the connection to hierarchical Bayesian estimation through the
shrinkage formula and a unified methodological progression, and distinguishes in-context
prediction from fine-tuning.
Section~\ref{sec:choice-structure} describes the two structural features of panel choice
data (set-valued occasions and persistent consumer heterogeneity) that motivate the
reformulations.
Section~\ref{sec:reformulation} introduces the proposed reformulation and the associated
data representations and task constructions.
Section~\ref{sec:empirical} describes the dataset, experimental design, and empirical
results.
Section~\ref{sec:conclusion} concludes.
Appendix~\ref{app:hb} provides a self-contained account of the HB framework.

\section{Related Literature}
\label{sec:lit}

We build on the discrete choice modeling tradition, including Bayesian methods for
consumer heterogeneity, and contribute to the literatures on machine learning for demand
estimation and TFMs. Across these streams, existing approaches either impose parametric
structure or require task-specific estimation on the focal dataset. We review each stream
in turn.

\paragraph{Discrete choice models.}
The discrete choice literature is grounded in the random utility framework, in which
observed choices arise from utility maximization within a choice set
\citep{mcfadden1974conditional}. Mixed logit extends the closed-form multinomial logit by
allowing coefficients to vary continuously across individuals, capturing unobserved
preference heterogeneity in panels of repeated choices \citep{revelt1998mixed};
\citet{mcfadden2000mixed} establish that the mixed logit family can approximate any random
utility model arbitrarily well, and \citet{train2009discrete} treats these models and their
simulation-based estimation. Latent class specifications offer a discrete alternative to
continuous heterogeneity \citep{kamakura1989probabilistic}, and \citet{berry1995automobile}
introduce a random-coefficients demand system estimated from market-level data.
\citet{chintagunta2011discrete} survey DCM applications in marketing;
\citet{berbeglia2022comparative} provide a systematic empirical comparison across DCM
families in retail settings.

Conjoint analysis is the principal stated-preference application of discrete choice, with
the methodological tradition originating in \citet{green1978conjoint} and formalized
within the random utility framework by \citet{louviere2000stated}. The Bayesian
heterogeneity estimation methods that dominate practice (reviewed below) were developed
largely in the conjoint setting but apply equally to revealed-preference panel data.

\paragraph{Preference heterogeneity and Bayesian methods.}
Accounting for consumer heterogeneity is central to both predictive accuracy and
managerial decision-making in marketing. \citet{lenk1996hierarchical} established the
hierarchical Bayes framework for conjoint analysis, showing that individual-level
partworth distributions can be recovered from designs too small to support individual
estimation by pooling observations through a common population prior.
\citet{rossi1996value} demonstrate the value of this approach for scanner panel data, and
\citet{allenby1998marketing} develop and compare hierarchical Bayes and finite mixture
specifications, providing the statistical framework that remains the benchmark for
individual-level preference recovery. \citet{rossi2005bayesian} codify this approach in a
comprehensive treatment of Bayesian methods for marketing DCMs.
\citet{dube2010state} use flexible heterogeneity models to disentangle structural state
dependence from spurious persistence driven by preference variation.
\citet{evgeniou2007convex} establish that the HB posterior mean is not uniquely Bayesian:
it coincides with penalized regression toward a population mean, with the population
prior precision playing the role of a regularization strength, placing HB within the
broader class of shrinkage estimators. Recent work extends the source of pooled information beyond the focal sample: \citet{ibragimov2026learning} transfer information across a firm's archive of past marketing campaigns to improve targeting in a focal campaign, and related approaches pool past experiments through observable campaign characteristics \citep{ellickson2023estimating, huang2024incrementality}. In these methods, the structure that supports transfer is estimated from the firm's own response data, and applying it requires fitting a model to the focal dataset.
Our heterogeneity encoding approaches, whether through a respondent identifier feature or
per-respondent in-context learning, pursue the same goal as HB (individual-level inference from short choice histories) but with a prior that is pretrained rather than parametrically estimated, applied by in-context learning rather than fitting. Section~\ref{sec:tfm-shrinkage} develops the connection between HB and TFMs. 

\paragraph{Demand estimation in operations management.}
DCMs are foundational to operations management, where choice
probabilities underlie assortment selection, pricing, and revenue management.
\citet{gallego2014nested} characterize optimal multi-product pricing under the
nested logit model, and \citet{farias2013nonparametric} show that nonparametric choice
models estimated from transaction data yield substantial revenue improvements over
parametric benchmarks in retail. Estimation from operational data introduces additional
challenges: sales records are censored by stockouts and assortment decisions, and
per-consumer purchase histories are short, with scanner panel data typically offering a
few dozen observations per individual. In this sparse-data regime, parametric Bayesian
shrinkage can distort estimates for atypical consumers as the population prior dominates
the likelihood; \citet{berbeglia2022comparative} document that DCM selection has
material consequences for operational decisions in retail.
Recent evidence suggests that the foundation model paradigm can ease these estimation
burdens in practice: \citet{sui2026bitter} find that a fine-tuned time-series foundation
model improves demand forecasting accuracy by 3.5--5.4\% over an incumbent production
system at Alibaba. Our results
contribute a parallel finding on the choice modeling side: TFMs offer an effective
alternative to HB, with the largest predictive advantage at 10--40 observations per
respondent, precisely the depth typical of scanner panels and revenue management
applications.

\paragraph{Machine learning methods for discrete choice.}
A growing literature examines the intersection of machine learning and discrete choice
modeling \citep{athey2019machine, cranenburgh2022choice}. One strand
develops hybrid architectures that embed neural networks within
utility-theoretic structure: \citet{sifringer2020enhancing} augment the multinomial logit
utility function with a learned representation component, \citet{wang2020deep}
demonstrate that deep neural networks can recover economically meaningful quantities such
as elasticities from choice data, and \citet{aouad2023rumnet} prove that any RUM can be
approximated arbitrarily well by a suitably constructed architecture. A second strand
forgoes structural constraints entirely: \citet{singh2023choice} exploit the
permutation-invariant structure of choice probabilities to construct flexible nonparametric
demand estimators, \citet{gabel2022product} develop a scalable deep learning model for
product choice with large assortments, and \citet{compiani2025demand} incorporate embeddings
from pretrained models into a mixed logit system to improve counterfactual demand
predictions. \citet{wang2023transformerchoicenet} develop a transformer that treats
assortment and purchase history as context, capturing cross-item interactions without a
prespecified utility structure.
All of these methods estimate task-specific parameters on the focal dataset. Our work is
distinguished in that the underlying model is held fixed at pretraining and applied purely
via in-context learning, with no task-specific parameters estimated. Rather than modifying the model, we focus on how data
representation and task formulation enable a pretrained row-wise architecture to capture
the relational and heterogeneous structure of choice data.

\paragraph{Large language models for preference elicitation.}
A nascent literature explores whether LLMs can serve as synthetic consumers.
\citet{horton2023homo} proposes treating LLMs as implicit computational models of human
agents, and \citet{chen2023rationality} demonstrate that GPT satisfies the generalized
axiom of revealed preference at higher rates than human subjects. Applied to
conjoint-style elicitation, LLMs recover willingness-to-pay estimates that are sometimes
comparable to human surveys but often inaccurate and occasionally wrong-signed
\citep{brand2023using}, and combining small human samples with LLM-generated data can
reduce estimation error \citep{wang2024augmentation}. Methodological challenges remain:
\citet{gui2023challenge} identify an identification problem when LLMs are blinded to
experimental design, and \citet{ludwig2024econometric} show that prompt and model choices
can yield dramatically different structural estimates absent a validation sample. Rather
than using language models as synthetic consumers, we apply TFMs to actual consumer choice
data, and we use in-context learning in its technical sense (labeled choice observations
as context from which the model predicts held-out choices) rather than eliciting
preferences through natural language prompting.

\paragraph{Tabular foundation models.}
Tabular machine learning has long been dominated by gradient-boosted decision trees
\citep{chen2016xgboost, ke2017lightgbm, prokhorenkova2018catboost}, which deep learning
has not consistently surpassed \citep{gorishniy2021revisiting, grinsztajn2022why,
shwartzziv2022tabular, erickson2025tabarena}. TFMs represent a qualitatively different
approach: pretrained offline on large collections of synthetic tasks and deployed via
in-context learning without task-specific parameter updates.

The theoretical foundation is the prior-data fitted network (PFN) framework of
\citet{muller2022transformers}, applied to tabular classification by
\citet{hollmann2023tabpfn} and extended to larger datasets by \citet{hollmann2025accurate};
\citet{nagler2023statistical} establishes its statistical consistency properties.
Section~\ref{sec:tfm-shrinkage} develops this framework in detail. Subsequent work has
scaled the approach \citep{priorlabs2025tabpfn25, qu2025tabicl} and addressed the
synthetic-to-real distribution gap \citep{ma2025tabdpt}. In-context learning in
transformers was established in the language domain \citep{brown2020language}, with
mechanistic accounts of the algorithms self-attention implements in context
\citep{akyurek2023what, vonoswald2023transformers}.

A key limitation of standard TFMs, including those above, is the assumption of row
independence: each observation is treated as an independent input--output pair during
inference. \citet{cucumides2026grables} formalize this limitation, proving that targets
driven by relational patterns across rows are inaccessible to any row-local predictor
regardless of model capacity. Discrete choice data are a specific and economically
motivated instance of this failure: choice probabilities depend on comparisons across
alternatives within a set, making them inaccessible to row-local models without
appropriate reformulation. Broader relational deep learning methods address inter-row
structure through graph neural networks over multi-table databases
\citep{fey2024relational, robinson2024relbench}, and \citet{eremeev2025turning} show that
pretrained TFMs can be repurposed as graph foundation models by treating node neighborhoods
as tabular context. Our approach addresses the relational structure of discrete choice data
not through graph construction but through structured data representations that encode
choice-set dependencies and individual heterogeneity within the existing row-wise TFM
framework.

\section{Tabular Foundation Models and Their Connection to Hierarchical Bayes}
\label{sec:tfm-shrinkage}
Hierarchical Bayesian DCMs, the dominant approach for individual-level
preference estimation in marketing, address sparse per-consumer data through a
specific mechanism: shrink individual estimates toward a population mean, with the
degree of pooling governed by how many choices each consumer has revealed. This
section introduces TFMs and develops their connection to
HB. The connection is more than analogy: both approaches solve the same sparse-data
problem through the same underlying principle (shrinkage toward a prior), differing
only in how that prior is specified. Understanding this connection clarifies what
TFMs add to the researcher's toolkit and what they require in exchange.

Section~\ref{subsec:pfn-icl} introduces in-context learning and the prior-data
fitted network (PFN) framework that underlies TabPFN.
Section~\ref{subsec:amortized-shrinkage} develops the HB-TFM connection formally,
presenting the HB shrinkage formula and positioning both approaches within the
broader methodological progression.
Section~\ref{subsec:icl-ft} distinguishes in-context prediction from fine-tuning
and explains the role of each in the empirical analysis.

\subsection{In-Context Learning and the PFN Framework}
\label{subsec:pfn-icl}

\paragraph{The in-context learning paradigm.}
The standard supervised-learning workflow trains a separate model on each new
dataset: specify a model family, optimize parameters on labeled observations,
tune hyperparameters via cross-validation, and evaluate on held-out data. This
procedure is repeated from scratch whenever the dataset or prediction target
changes. \emph{In-context learning} (ICL) replaces per-dataset parameter fitting
with a conditioning operation. A single network $q_\phi$ is pretrained once across
a large collection of tasks. At prediction time, $q_\phi$ receives the labeled
training set $\mathcal{D} = \{(x_i, y_i)\}_{i=1}^n$ as part of its input---the
``context''---and produces a prediction for a new point $x_*$ in a single forward
pass, with the weights $\phi$ held fixed:
\begin{equation}
  \hat{y}_* = q_\phi(x_*, \mathcal{D}).
  \label{eq:icl}
\end{equation}
The analogy to large language models is instructive: when a user prompts ChatGPT
with several input-output examples followed by a new input, the model produces a
prediction without any weight update \citep{brown2020language}. TabPFN performs
the same operation on rows of a table---the labeled training rows are the context,
and the test row is the new input---exploiting the same underlying mechanism,
self-attention over context tokens, adapted to tabular data.

\paragraph{The prior-data fitted network framework.}
\citet{muller2022transformers} provide a Bayesian interpretation of in-context
learning through the prior-data fitted network (PFN) framework. Given a training
set $\mathcal{D}$ and a model family parameterized by $\theta$, the Bayesian
posterior predictive distribution for a test outcome $y_*$ at input $x_*$ is
\begin{equation}
  p(y_* \mid x_*, \mathcal{D})
  =
  \int
    p(y_* \mid x_*, \theta)\,
    p(\theta \mid \mathcal{D})\,
    d\theta.
  \label{eq:pfn_predictive}
\end{equation}
Equation~\eqref{eq:pfn_predictive} averages the likelihood over all parameter
values weighted by their posterior given the training data, simultaneously
accounting for model fit and parameter uncertainty. For most model families this
integral is analytically intractable and requires MCMC, which is expensive to run
anew for each new dataset.

The PFN insight is to amortize this computation. Rather than evaluating
\eqref{eq:pfn_predictive} at prediction time, the network $q_\phi$ is trained
once to approximate it. Pretraining proceeds by repeatedly (i)~sampling a
complete synthetic dataset from a prior $p(\mathcal{D})$ over data-generating
processes, (ii)~splitting it into a context $\mathcal{D}$ and held-out test
pairs $\{(x_*^{(k)}, y_*^{(k)})\}$, and (iii)~updating $\phi$ to minimize the
negative log-predictive loss on the held-out pairs given the context. After
pretraining, $q_\phi(y_* \mid x_*, \mathcal{D}) \approx p(y_* \mid x_*, \mathcal{D})$ for any dataset drawn from the prior, in a single forward pass. The weights
$\phi$ parameterize a learned prediction algorithm; they do not represent an
explicit posterior over $\theta$, but rather a general prediction procedure
trained to behave like a Bayesian predictor across the full distribution of
pretraining tasks.

\paragraph{The pretraining prior.}
The prior $p(\mathcal{D})$ determines which prediction tasks the network handles
well. TabPFN's prior combines two families of synthetic data-generating processes:
structural causal models, which generate datasets with realistic correlation
structure and causal directionality, and Bayesian neural networks with randomly
drawn weights, which produce a rich family of smooth and nonsmooth response
surfaces. Simpler data-generating processes are up-weighted, encoding an Occam's
razor preference for regular relationships. We refer to the statistical
regularities that $\phi$ encodes through pretraining on this prior as the TFM's
\emph{learned prior}: the general prediction heuristics internalized from millions
of synthetic tasks. Before fine-tuning, this learned prior is general-purpose---not
estimated from the focal choice panel, not calibrated to the target product
category, and not adapted to the specific data under study.

\paragraph{Statistical properties.}
\citet{hollmann2023tabpfn} demonstrated that this framework is practically
viable, showing that a single forward pass matches or exceeds tuned AutoML
pipelines on small classification benchmarks. \citet{hollmann2025accurate}
scaled the approach to datasets with up to 10,000 training rows and 500 features,
achieving state-of-the-art accuracy while completing inference in seconds.
\citet{nagler2023statistical} provides the first frequentist analysis of these
predictors, and two properties bear on our argument. The prediction variance
declines as the context size $n$ grows, at a rate that depends on how strongly
the network responds to any single context example rather than at a universal
$1/n$ rate. Low variance does not, however, guarantee accuracy: the predictor is
consistent only when the pretraining prior $p(\mathcal{D})$ places sufficient mass
on the true data-generating process, a condition that the architecture controlling
variance does not itself ensure. A model pretrained on generic synthetic data may be biased
on choice data whose structure lies outside the pretraining distribution; this
observation motivates the reformulation developed in
Sections~\ref{sec:choice-structure} and \ref{sec:reformulation}.

\subsection{HB and TFMs: A Unified Shrinkage View}
\label{subsec:amortized-shrinkage}

\paragraph{The HB model.}
The HB approach resolves the sparse-data problem by treating individual preference
parameters as draws from a common population distribution \citep{rossi2005bayesian,
allenby1998marketing}. Let $\boldsymbol{\beta}_i \in \mathbb{R}^K$ denote consumer
$i$'s vector of taste parameters. At the population level, $\boldsymbol{\beta}_i \sim \mathcal{N}(\boldsymbol{\mu}, \boldsymbol{\Sigma})$, where $\boldsymbol{\mu} \in \mathbb{R}^K$ is the population mean taste vector and
$\boldsymbol{\Sigma}$ is the population covariance matrix, both estimated jointly
from the full panel. At the individual level, consumer $i$'s choices are governed
by a likelihood $p(y_{it} \mid \mathbf{x}_{it}, \boldsymbol{\beta}_i)$, where
$\mathbf{x}_{it}$ collects observed alternative attributes on choice occasion $t$.
Appendix~\ref{app:hb} develops the full MNL specification used in the empirical
analysis.

\paragraph{The HB shrinkage formula.}
The key shrinkage property is most transparent under the conjugate specification.
Let $\mathbf{X}_i$ denote the $T_i \times K$ matrix stacking consumer $i$'s
attribute vectors across $T_i$ choice occasions. Under the Gaussian individual
likelihood $y_{it} = \mathbf{x}_{it}'\boldsymbol{\beta}_i + \varepsilon_{it}$
with $\varepsilon_{it} \stackrel{\mathrm{iid}}{\sim} \mathcal{N}(0, \sigma^2)$,
the model is conjugate and the posterior mean of $\boldsymbol{\beta}_i$ has the
closed form \citep{lindley1972bayes, rossi2005bayesian}:
\begin{equation}
\bar{\boldsymbol{\beta}}_i
\;=\;
(\mathbf{I} - \mathbf{B}_i)\,\hat{\boldsymbol{\beta}}_i^{\,\mathrm{MLE}}
\;+\;
\mathbf{B}_i\,\boldsymbol{\mu},
\label{eq:hb_shrinkage}
\end{equation}
where $\mathbf{B}_i = \bigl(\mathbf{I} +
\boldsymbol{\Sigma}\mathbf{X}_i'\mathbf{X}_i/\sigma^2\bigr)^{-1}$ is the shrinkage
matrix and $\hat{\boldsymbol{\beta}}_i^{\,\mathrm{MLE}}$ is consumer $i$'s
maximum likelihood estimate computed from $i$'s data alone. For scanner panel data
and choice-based conjoint, the logit likelihood replaces the Gaussian, breaking
conjugacy; equation~\eqref{eq:hb_shrinkage} holds approximately, and the qualitative
behavior is the same. The posterior mean is a matrix-weighted average of the
consumer's own data and the population mean $\boldsymbol{\mu}$, governed by the
shrinkage matrix $\mathbf{B}_i$. When consumer $i$ has many observations,
$\mathbf{X}_i'\mathbf{X}_i$ is large, $\mathbf{B}_i \to \mathbf{0}$, and
$\bar{\boldsymbol{\beta}}_i$ approaches the individual MLE. When observations are
sparse, $\mathbf{B}_i \to \mathbf{I}$ and $\bar{\boldsymbol{\beta}}_i$ collapses
to $\boldsymbol{\mu}$. This adaptive pooling is the mechanism that makes HB produce
stable individual-level estimates even from short purchase histories.

\paragraph{Shrinkage and regularization.}
\citet{evgeniou2007convex} show that the HB posterior mean is not uniquely Bayesian:
the shrinkage formula \eqref{eq:hb_shrinkage} is equivalent to penalized regression
toward a population mean, with $\boldsymbol{\Sigma}^{-1}$ playing the role of a
regularization strength. HB is therefore an instance of a broader principle: shrinkage
toward a shared representation, which also underlies the methods reviewed below.

\paragraph{The unified progression.}
Every method for individual-level preference estimation from sparse panel data is a
shrinkage estimator: it pools information across consumers toward a shared
representation, with the degree of pooling governed by the available data. Methods
differ in how that shared prior is specified. Table~\ref{tab:shrinkage} organizes
this progression.

\begin{table}[t]
\centering
\caption{Methods for individual-level preference recovery as shrinkage estimators,
organized by how the shared prior is specified.}
\label{tab:shrinkage}
\small
\begin{tabular}{p{3.0cm}p{5.8cm}p{4.2cm}}
\hline\hline
\textbf{Method} & \textbf{How the shared prior is specified} & \textbf{Key references} \\
\hline
Individual OLS &
No pooling; consumers estimated independently &
\citet{green1978conjoint} \\[4pt]
Latent class &
Discrete mixture of $K$ segments; EM-estimated from focal sample &
\citet{kamakura1989probabilistic} \\[4pt]
Hierarchical Bayes &
Continuous $\mathcal{N}(\boldsymbol{\mu},\boldsymbol{\Sigma})$;
MCMC-estimated from focal sample &
\citet{lenk1996hierarchical}; \citet{allenby1998marketing} \\[4pt]
Ridge / multi-task &
Implicit Gaussian; regularization parameter by cross-validation &
\citet{evgeniou2007convex} \\[4pt]
TFM &
Learned from synthetic datasets via pretraining; applied via ICL &
\citet{muller2022transformers}; \citet{hollmann2025accurate} \\
\hline\hline
\end{tabular}
\end{table}

TFMs occupy the rightmost position in this progression. In HB, the prior
$\mathcal{N}(\boldsymbol{\mu},\boldsymbol{\Sigma})$ is a parametric family
specified by the analyst and estimated from the focal sample. In a TFM, the prior
is encoded in the pretrained weights $\phi$, learned from millions of synthetic
datasets outside the focal sample. At prediction time, each consumer's in-context
examples condition the pretrained prior on that consumer's observed choices: the
functional analog of the HB posterior update in \eqref{eq:hb_shrinkage}, but
computed in a single forward pass without MCMC.

Three concrete differences follow. First, computational speed: a single forward pass
replaces MCMC, reducing inference from minutes to seconds. Second, distributional
flexibility: the learned prior is not restricted to the multivariate normal assumed
by standard HB and can encode irregular distributions, nonlinear feature
interactions, and response patterns that the normal prior would smooth over. Third,
cold-start performance: because the prior is estimated from synthetic tasks rather
than the focal sample, a TFM can produce calibrated predictions even from a single
choice observation per consumer.

In exchange, the learned prior is not directly interpretable as a distribution over
preferences: the practitioner cannot inspect $\boldsymbol{\mu}$ and
$\boldsymbol{\Sigma}$ or assess the direction of shrinkage for a given consumer.
There are no posterior draws, so formal uncertainty quantification requires different
tools than the standard HB output. And the learned prior may be biased if the true
data-generating process falls outside the pretraining distribution, the risk that
\citet{nagler2023statistical} formalizes and that the reformulations in
Sections~\ref{sec:choice-structure} and \ref{sec:reformulation} are designed to
address.

\subsection{In-Context Prediction and Fine-Tuning}
\label{subsec:icl-ft}

Given the learned prior encoded in the pretrained weights, there are two ways to
apply a TFM to a specific choice panel: in-context prediction, which keeps the
weights fixed, and fine-tuning, which adapts those weights before prediction.

In-context prediction keeps the pretrained weights $\phi$ fixed. Given a context
$\mathcal{D}$ and test covariates $x_*$, prediction is formed as $q_{\phi}(x_*, \mathcal{D})$. No gradient updates are performed on the focal dataset. Adaptation occurs only
through the examples supplied in context and their representation. For panel choice
data, population information can therefore enter only through the organization of
the context: by pooling consumers in a shared table, including consumer identity
as a feature, or constructing separate per-consumer contexts.

Fine-tuning changes the model weights before in-context prediction. Starting from
the pretrained weights $\phi$, the model is further trained on pooled choice
observations from the focal population: $\phi \xrightarrow{\text{fine-tune on }\mathcal{D}_{\mathrm{pop}}} \phi_{\mathrm{choice}}$. Operationally, this is gradient-based adaptation of neural network weights. It is
not Bayesian updating: the procedure does not analytically condition a prior or
produce posterior draws over preference parameters. Conceptually, it shifts the
learned prior from the off-the-shelf model, which encodes general tabular
regularities, toward one adapted to the empirical regularities of the focal choice
population. After fine-tuning, predictions use the adapted weights
$q_{\phi_{\mathrm{choice}}}(x_*, \mathcal{D}_i)$, where $\mathcal{D}_i$ denotes the consumer-specific
calibration context supplied at prediction time. Fine-tuning does not replace
in-context prediction; it changes the pretrained mapping through which the context
is interpreted.

\section{From Row-Level Prediction to Panel Choice Data}
\label{sec:choice-structure}

Panel choice data have structure that row-and-column prediction does not natively
represent. A purchase occasion is not a single labeled row, but a choice among
alternatives in a set; and repeated choices by the same consumer are linked by
persistent preferences. This section examines both features and, in
Section~\ref{subsec:implications-reformulation}, draws out their implications for
applying a learned prior to panel choice prediction. The goal is to identify what
information must be made visible to the model.

\subsection{Set-Valued Choice Occasions}
\label{subsec:set-valued-choice}

DCMs describe consumer decisions through random utility maximization: a consumer selects
the alternative yielding the greatest utility, decomposed into a component explained by
observed attributes and an unobserved stochastic term. Consumer $i$ derives utility
\begin{equation}
    u_{ijt} = \mathbf{x}_{ijt}'\boldsymbol{\beta}_i + \varepsilon_{ijt}
    \label{eq:utility}
\end{equation}
from alternative $j$ in choice set $S_t$ on occasion $t$, where
$\mathbf{x}_{ijt}$ are observed alternative attributes, $\boldsymbol{\beta}_i$ is a
consumer-specific vector of taste parameters, and $\varepsilon_{ijt}$ is an idiosyncratic
random shock. Under i.i.d.\ Type~I extreme-value errors, the probability that consumer \(i\) chooses
alternative \(j\) from choice set \(S_t\) on occasion \(t\) takes the
multinomial logit form
\begin{equation}
    P_{ijt}
    =
    \frac{\exp(\mathbf{x}_{ijt}'\boldsymbol{\beta}_i)}
         {\sum_{k \in S_t}
         \exp(\mathbf{x}_{ikt}'\boldsymbol{\beta}_i)} .
    \label{eq:mnl}
\end{equation}
The specification \eqref{eq:mnl} has two features. First, the prediction object is a distribution over a set:
the probability assigned to one alternative is defined relative to the
other alternatives available on the same occasion. Second, the
alternatives in \(S_t\) are unordered. A model's prediction should not
depend on whether the chosen alternative is listed first, second, or fourth in
the data representation.

These features differ from the native input-output structure of standard
tabular prediction. In a row-level supervised task, each observation is a
feature vector paired with a label. In a choice task, the observed outcome
is attached to a group of rows: exactly one alternative in the set is
chosen and the remaining alternatives are not. If each alternative is
converted into an independent row, the model may see the attributes of the
chosen and unchosen alternatives, but the mutually exclusive nature of the
choice set is no longer represented explicitly. The set-valued structure
therefore motivates reformulations that make comparison within a choice
occasion more visible, such as wide representations, pairwise comparisons,
within-set rank features, or post-processing that normalizes predicted
probabilities within each set.


\subsection{Repeated Choices and Consumer Heterogeneity}
\label{subsec:consumer-heterogeneity}

Panel choice data contain repeated observations from the same decision
maker. In the random utility notation above, the same
\(\boldsymbol{\beta}_i\) governs all observed choices by consumer \(i\).
Thus, a consumer's earlier purchases are informative about their later
purchases because they reveal persistent features of that consumer's
preferences. This cross-observation dependence is the reason that
heterogeneity has played such a central role in marketing DCMs:
hierarchical Bayesian, finite-mixture, and latent-class specifications all
aim to recover consumer-level preference differences while borrowing
strength across consumers
\citep{allenby1998marketing,rossi2005bayesian,train2009discrete}.

A TFM does not automatically represent this
structure. TabPFN can attend to all examples supplied in context, but the
model must be given a representation that tells it which observations
belong to the same consumer, or it must be organized so that a
consumer's own history forms the relevant context for prediction.
Without such information, repeated choices from the same consumer are
simply pooled examples in a supervised prediction task. The model may
still learn average relationships between features and choices, but it
has no direct way to personalize predictions using the consumer's own
purchase history.

This distinction maps directly onto the unified shrinkage view developed in
Section~\ref{sec:tfm-shrinkage}. In HB-MNL, population information enters through an
estimated distribution over consumer taste parameters, and individual-level predictions
shrink toward that distribution when consumer histories are sparse. In TabPFN, population
and consumer information enter differently. With fixed pretrained weights, the learned
prior is unchanged; the model can use population or consumer structure only to the extent
that the context organization exposes it. For example, including consumer identity as a
categorical feature keeps all consumers in a shared context while
marking repeated observations from the same consumer. Constructing one
context per consumer instead emphasizes personalization from that
consumer's own history but reduces direct borrowing across consumers
within the in-context task.


\subsection{Implications for TabPFN Reformulation}
\label{subsec:implications-reformulation}

The preceding discussion suggests two axes for adapting TabPFN to panel
choice data. The first is the \emph{choice-representation axis}: how to
convert a set-valued choice occasion into inputs suitable for a
row-and-column prediction model. A long representation treats each
alternative as a row; a wide representation stores the entire choice set
in one row; a pairwise representation converts each observed choice into
comparisons between alternatives; and rank or set-summary features make
relative position within the choice set more explicit.

The second is the \emph{heterogeneity axis}: how to make repeated
consumer histories available to the model. A pooled representation
suppresses consumer identity and asks the model to learn average choice
patterns. A respondent-categorical representation includes consumer
identity as a feature, allowing the model to condition on repeated
observations from the same consumer within a shared population context.
A per-respondent in-context representation treats each consumer as its
own prediction task, using that consumer's calibration choices as the
context for its future choices. Fine-tuning, studied separately in
Section~\ref{subsec:results-ft}, adds another route by updating the
pretrained model on pooled choice data before prediction.

These two axes are conceptually distinct. Choice representation determines how the model
expresses the within-occasion comparison; heterogeneity encoding determines how the model
can use information across occasions and across consumers. Section~\ref{sec:reformulation} operationalizes
these axes as concrete TabPFN inputs. Section~\ref{sec:empirical} then
evaluates them empirically, distinguishing gains that come from
representing the choice set differently from gains that come from making
consumer heterogeneity visible to the model.

\section{Reformulating Panel Choice Data for TabPFN}
\label{sec:reformulation}

Section~\ref{sec:choice-structure} identified two design axes for adapting
TabPFN to panel choice data. The first is how to represent a set-valued
choice occasion as tabular inputs. The second is how to expose consumer
heterogeneity to the model. This section describes the concrete
reformulations we evaluate. We organize the discussion around these two
axes rather than around individual specifications, because the empirical
analysis asks which source of information accounts for predictive gains:
choice-set representation or consumer heterogeneity.
Sections~\ref{subsec:choice-representations} and \ref{subsec:relative-set-features}
develop the representation axis and Section~\ref{subsec:heterogeneity-encodings} the
heterogeneity axis; Section~\ref{subsec:method-icl-ft} then contrasts in-context with
fine-tuned prediction, a further dimension along which these reformulations can be applied.

\subsection{Choice-Set Representations}
\label{subsec:choice-representations}

Let \(S_t=\{1,\ldots,J_t\}\) denote the alternatives available on choice
occasion \(t\), and let \(y_{ijt}=1\) if consumer \(i\) chooses
alternative \(j\) on that occasion. Each alternative has attributes
\(\mathbf{x}_{ijt}\). A TabPFN classifier requires tabular rows and
labels, so the first step is to convert the choice occasion into a
supervised prediction problem. We consider three main representations:
long, wide, and pairwise.

\paragraph{Long representation.}

The long representation treats each alternative in a choice set as one
row. For consumer \(i\), occasion \(t\), and alternative \(j\), the
input row is \(\mathbf{x}_{ijt}\), and the label is
\[
    y_{ijt} =
    \begin{cases}
    1, & \text{if alternative } j \text{ is chosen on occasion } t, \\
    0, & \text{otherwise.}
    \end{cases}
\]
Thus, a four-alternative choice occasion produces four binary rows. This
representation is closest to the row-level classification format for
which TabPFN is designed. It also preserves alternative-level attributes
directly. However, the choice-set constraint is not automatic: the model
produces a binary score for each alternative row, and we convert these
scores into choice probabilities by normalizing them within the choice
set.

Specifically, let \(\hat{s}_{ijt}\) denote the TabPFN score or predicted
probability assigned to alternative \(j\). We form choice probabilities
as
\[
    \hat{P}_{ijt}
    =
    \frac{\hat{s}_{ijt}}
    {\sum_{k \in S_t} \hat{s}_{ikt}} .
\]
This post-processing step restores the requirement that predicted
probabilities sum to one within each choice set.

\paragraph{Wide representation.}

The wide representation stores the entire choice set in a single row.
The attributes of all alternatives in \(S_t\) are concatenated into one
feature vector,
\[
    \mathbf{z}_{it}
    =
    \left(
    \mathbf{x}_{i1t},
    \mathbf{x}_{i2t},
    \ldots,
    \mathbf{x}_{iJ_t t}
    \right),
\]
and the label is the identity of the chosen alternative. Therefore, the wide
representation becomes a multi-class classification problem.

The wide representation makes the mutually exclusive nature of the
choice occasion explicit: one row corresponds to one choice, and the
classifier directly predicts the chosen alternative. Its primary
limitation is that it does not encode the repeated-alternative
structure of the problem: the TFM observes separate columns for each
alternative's attributes but is not told that corresponding columns
across alternatives represent the same underlying feature.
Relationships such as price comparisons across brands must therefore
be inferred from the data rather than supplied by the representation.
The wide representation also requires a fixed alternative ordering; we
consider permutation augmentation to reduce dependence on arbitrary
column positions.

\paragraph{Pairwise representation.}

The pairwise representation converts a multinomial choice into binary
comparisons. For each observed choice of alternative \(j\) over another
available alternative \(k\), we construct a row describing the
comparison between the two alternatives. In the difference version, the
input is $\mathbf{x}_{ijt} - \mathbf{x}_{ikt}$, and the label indicates that \(j\) is preferred to \(k\). In the ordered
version, the input concatenates the two alternatives, $[\mathbf{x}_{ijt}, \mathbf{x}_{ikt}]$, and the label indicates whether the first alternative is the one chosen.

Pairwise representations are attractive because they express choice as a
set of revealed preference inequalities. However, they also change the
prediction problem: the model learns binary comparisons rather than
direct choice probabilities over the full set. To evaluate a choice
occasion, we aggregate pairwise predictions back to alternative-level
scores and then normalize those scores within the choice set. We
consider several aggregation rules in the empirical sweep, including
Borda-style aggregation and logit-sum aggregation.

\subsection{Relative and Set-Level Features}
\label{subsec:relative-set-features}

The long representation is row-natural but does not by itself tell the
model how an alternative compares with the other alternatives available
on the same occasion. We therefore consider simple features that make
within-set relative position explicit.

The first set of features are rank features. For each attribute that
varies within a choice set, we compute the rank of an alternative
relative to the other alternatives in that set. In the yogurt
application considered in our numerical study, price varies by brand and occasion, so a price-rank feature
indicates whether a brand is relatively cheap or expensive within the
current choice set. Rank features are attractive because they are
scale-free and directly encode comparison, while keeping the long
representation in a row-level format.

The second set of features are set-summary features. These features
describe the choice set around an alternative, such as within-set
means, minima, maxima, or deviations from the set average. The goal is
to give the model information about the competitive environment faced
by the focal alternative. Such features provide the row-level classifier with
information that would otherwise be implicit in the full choice set.

These features are useful for separating two questions empirically. If
relative or set-summary features improve prediction substantially, then
making within-set comparison more explicit is important. If their
effect is small relative to the effect of consumer heterogeneity, then
the main challenge lies less in representing the choice set and more
in representing persistent consumer preferences.

\subsection{Encoding Consumer Heterogeneity}
\label{subsec:heterogeneity-encodings}

The second design axis concerns consumer heterogeneity. We compare
three ways of organizing the same panel data: pooled prediction,
respondent-categorical prediction, and per-respondent in-context
prediction.

\paragraph{Pooled prediction.}

The pooled specification suppresses consumer identity. Training rows
from all consumers are combined into a single context, but the model
is not told which rows belong to the same consumer. This specification
is the closest TabPFN analogue to a homogeneous DCM. It allows
the model to learn average relationships between alternative attributes
and choices, but it does not provide a direct channel for
consumer-specific preferences.

\paragraph{Respondent-categorical prediction.}

The respondent-categorical specification includes consumer identity as
a categorical feature. All consumers remain in a shared context, but
repeated observations from the same consumer share the same respondent
identifier. This construction gives the model two kinds of information
at once. First, because all consumers are pooled into one context, the
model can condition on population-level regularities in the data.
Second, because consumer identity is included as a feature, the model
can use a consumer's own past choices to personalize predictions for
that consumer.

This construction is not equivalent to estimating a random coefficient
vector for each consumer. The respondent identifier is a feature
supplied to the pretrained model, and the model decides how to use it
through its learned in-context prediction mechanism. Nevertheless, it
is the most direct way to make repeated consumer histories visible
while retaining a shared population context.

\paragraph{Per-respondent in-context prediction.}

The per-respondent specification creates a separate ICL
problem for each consumer. For consumer \(i\), the context consists
only of that consumer's calibration choices, and the test rows are
that same consumer's holdout choices. This construction emphasizes
personalization: the model predicts future choices for a consumer
using only that consumer's own observed history.

The advantage of this approach is that it aligns naturally with the
panel prediction problem for an existing consumer. The limitation is
that it removes direct cross-consumer pooling from the in-context
task. When a consumer has few calibration choices, the context may be
too small for the model to infer stable consumer-specific patterns.
This limitation is especially important in shallow panels and motivates
the fine-tuning analysis in Section~\ref{subsec:results-ft}, where
population information can enter through weight updates before
consumer-level prediction.

\subsection{In-Context Prediction and Fine-Tuned Prediction}
\label{subsec:method-icl-ft}

All reformulations can be used with TabPFN in-context prediction. In
this case, the pretrained weights remain fixed, and the model adapts
to the choice task only through the context supplied at prediction
time. This is the setting evaluated in
Section~\ref{subsec:results-icl}. It asks how much the TFM's learned prior can do when the choice data
are represented in different ways and when consumer heterogeneity is or
is not made visible to the model.

We also evaluate fine-tuned TabPFN for a subset of representative
specifications. Fine-tuning starts from pretrained TabPFN weights and
updates them using the pooled training observations from the focal
choice population. After fine-tuning, the model is evaluated using the
same train-holdout split and the same prediction metrics as in the
in-context analysis. As discussed in Section~\ref{subsec:icl-ft}, we
interpret fine-tuning as moving the model from the TFM's learned prior
toward a population choice-aware prior. The fine-tuning
analysis is therefore not a separate method family, but a second way
of using TabPFN: instead of only conditioning a fixed prior on
context, the model first adapts its weights to the empirical
regularities of the target choice population.

\section{Empirical Application}
\label{sec:empirical}

We evaluate the proposed reformulations using a yogurt scanner panel, a
revealed-preference dataset in which consumers make repeated brand choices
across grocery shopping occasions. We first describe the data, evaluation
protocol, and benchmark DCMs, then evaluate TabPFN with fixed pretrained
weights, and finally evaluate fine-tuned TabPFN. Together, these analyses show
when a tabular foundation model can compete with hierarchical DCMs, and which
parts of the gain come from representation, heterogeneity, and
population-specific prior updating.

\subsection{Data and Benchmarks}
\label{subsec:data-benchmarks}

We use the yogurt scanner panel originally analyzed by
\citet{jain1994random} and distributed in the \texttt{logitr} package in R.
The data contain repeated grocery purchase decisions from 98 consumers
(those with at least five purchase occasions in the raw data), each
choosing among four yogurt brands: Dannon, Hiland, Weight Watchers, and
Yoplait. There is no outside option, so every observed occasion is a
purchase of one of the four brands. Each brand corresponds to one
alternative in the four-element choice set and is described by five
features: three brand indicators, a price, and a
binary indicator for whether the brand was featured in a newspaper
advertisement. This setting is well suited to our analysis because it
is a revealed-preference panel: consumers make actual purchase
decisions under prices and promotional conditions that are not designed
by the researcher, and the same consumers are observed repeatedly over
time.

The number of purchase occasions varies substantially across
consumers. This variation is central to our empirical design because
it allows us to study how each method performs as the amount of
consumer-level choice history changes. We therefore group consumers
into four purchase-depth bins, shown in Table~\ref{tab:depth_bins}. The
shallow bin contains consumers for which individual preferences are
weakly identified without pooling, whereas the deep bin contains
consumers with longer histories that can support more reliable
individual-level inference. Thus, consumer depth is not merely a
feature of the data, but an empirical axis for studying when different
forms of shrinkage and heterogeneity modeling are most useful.

\begin{table}[h!]
\centering
\caption{Consumer distribution by purchase depth.}
\label{tab:depth_bins}
\small
\begin{tabular}{lccc}
\hline\hline
\textbf{Bin} & \textbf{Occasions per consumer}
             & \textbf{Consumers} & \textbf{Description} \\
\hline
B1 & 5--10  & 31 & Shallow \\
B2 & 11--20 & 33 & Medium \\
B3 & 21--40 & 21 & Medium-deep \\
B4 & $>$40  & 13 & Deep \\
\hline
Total & --- & 98 & \\
\hline\hline
\end{tabular}
\end{table}

For each consumer, we reserve the three most recent purchase occasions
as holdout observations and use all earlier occasions for estimation or
in-context prediction. This protocol is held fixed across all methods,
so every model is evaluated on the same future choice occasions. The
training observations serve as estimation data for the benchmark choice
models. For TabPFN, they serve as in-context examples: in the
respondent-categorical construction, observations from all consumers
are pooled into a shared context with consumer identity included as a
categorical feature; in the per-respondent construction, each
consumer's training observations form a separate context for predicting
that consumer's holdout choices. This design matches the
panel-prediction problem faced by firms that use past purchase
histories to predict subsequent choices by existing customers.

We evaluate predictive performance using two metrics. The primary
metric is holdout log-likelihood (LL), defined as the average
per-occasion log predicted probability assigned to the chosen brand on
held-out occasions. Higher values indicate better probabilistic choice
prediction. We also report hit rate, defined as the fraction of
held-out occasions for which the model assigns the highest predicted
probability to the brand actually chosen. Log-likelihood evaluates the
full predicted choice distribution, while hit rate summarizes argmax
accuracy. When we express differences in these metrics as percentages,
log-likelihood differences are reported as $\exp(\Delta\mathrm{LL})-1$,
the implied proportional change in the per-occasion predicted
probability of the chosen brand, and hit-rate differences are reported
relative to the benchmark hit rate.

We compare TabPFN against four discrete-choice benchmarks. The first is
pooled MNL, which imposes homogeneous preferences across consumers and
serves as the classical non-hierarchical baseline. The remaining three
are hierarchical Bayesian mixed-logit models estimated using
\texttt{bayesm} \citep{rossi2005bayesian}. The first HB specification
assumes that consumer-level taste parameters are drawn from a single
multivariate normal population distribution. The second allows the
population distribution to be a finite mixture of normals, which can
capture discrete segments or multimodal heterogeneity. The third uses a
Dirichlet process prior, allowing the number and composition of latent
preference groups to be learned more flexibly from the data. Together,
these benchmarks span the standard homogeneous model, the canonical HB
specification widely used in marketing practice, and more flexible
heterogeneous specifications. For the Bayesian benchmarks, holdout
choice probabilities are computed by averaging choice probabilities
over retained posterior draws, rather than by evaluating the choice
model only at posterior mean parameters.

We use TabPFN \citep{hollmann2025accurate} as the representative TFM. The ICL
results use TabPFN v2.6, the strongest released version available for
off-the-shelf prediction in our experiments; the fine-tuning results use TabPFN
v2.5, for the reason given in Section~\ref{subsec:results-ft}.

\subsection{TabPFN In-Context Learning Results}
\label{subsec:results-icl}

We first evaluate TabPFN in its ICL form, keeping the
pretrained model weights fixed. In this analysis, TabPFN brings to the
yogurt task a learned prior from pretraining,
but that prior is not updated using the yogurt choice data. Any
adaptation to the focal market therefore occurs through the calibration
observations supplied in context and through the way those observations
are represented. This setting allows us to separate two sources of
predictive improvement: gains from reformulating the choice problem for
a TFM, and gains from exposing consumer-level
heterogeneity to the model.

We begin with the homogeneous setting, where all consumers are pooled
and the model is not given any consumer identifier or
consumer-specific context. Table~\ref{tab:icl_homogeneous} compares
pooled MNL with selected homogeneous TabPFN reformulations. The full
sweep contains many additional representation variants; here we report
representative specifications that span the main representation
families.

\begin{table}[h!]
\centering
\caption{Homogeneous in-context prediction results.}
\label{tab:icl_homogeneous}
\small
\begin{tabular}{lrr}
\hline\hline
\textbf{Method} & \textbf{LL} & \textbf{Hit} \\
\hline
Pooled MNL                         & $-1.153$ & $0.476$ \\
\hline
TabPFN, wide                       & $-1.149$ & $0.463$ \\
TabPFN, long                       & $-1.115$ & $0.483$ \\
TabPFN, long with rank features    & $-1.131$ & $0.452$ \\
TabPFN, pairwise                   & $-1.151$ & $0.476$ \\
\hline\hline
\end{tabular}
\end{table}

The homogeneous results show that changing the representation alone
yields only modest improvements. The best homogeneous TabPFN
specification, the long representation, improves predicted probability
by approximately $3.9\%$ over pooled MNL, but the gain is small
relative to the improvements reported below once consumer
heterogeneity is introduced. Other representations are close to pooled
MNL, and some improve one metric while worsening the other. This
pattern suggests that the learned prior, by itself,
does not solve the main statistical problem in the panel. The dominant
source of predictive difficulty is not simply the row-wise format of
the data, but the fact that repeated observations from the same
consumer reflect persistent consumer-level preferences.

We next introduce consumer heterogeneity into the TabPFN context. We
consider two ways of doing so. In the respondent-categorical
construction, all consumers remain in a shared context and consumer
identity is included as a categorical feature. In the per-respondent
construction, each consumer forms a separate ICL
problem, using that consumer's calibration choices to predict its
holdout choices. These two constructions differ in how population
information can enter the prediction problem, but they share the same
purpose: both expose consumer-level heterogeneity to the model.

Table~\ref{tab:icl_heterogeneous} compares the heterogeneous TabPFN
specifications with the benchmark DCMs. We evaluated 37 TabPFN variants
across the long, wide, and pairwise reformulation families; the table
reports a representative selection spanning these families.

\begin{table}[h!]
\centering
\caption{Heterogeneous in-context prediction results.}
\label{tab:icl_heterogeneous}
\small
\begin{tabular}{llrr}
\hline\hline
\textbf{Family} & \textbf{Method} & \textbf{LL} & \textbf{Hit} \\
\hline
\multicolumn{4}{l}{\emph{Benchmarks}} \\
\quad Homogeneous benchmark
    & Pooled MNL
    & $-1.153$ & $0.476$ \\
\quad Hierarchical benchmark
    & HB-MNL, single normal population
    & $-0.725$ & $0.745$ \\
\quad Hierarchical benchmark
    & HB-MNL, finite mixture
    & $-0.735$ & $0.735$ \\
\quad Hierarchical benchmark
    & HB-MNL, Dirichlet process prior
    & $-0.741$ & $0.735$ \\
\hline
\multicolumn{4}{l}{\emph{Long TabPFN}} \\
\quad Respondent-categorical
    & Long
    & $-0.671$ & $0.762$ \\
\quad Per-respondent ICL
    & Long
    & $-0.725$ & $0.765$ \\
\quad Respondent-categorical
    & Long with rank features
    & $\mathbf{-0.648}$ & $\mathbf{0.772}$ \\
\quad Per-respondent ICL
    & Long with rank features
    & $-0.732$ & $0.752$ \\
\hline
\multicolumn{4}{l}{\emph{Wide TabPFN}} \\
\quad Respondent-categorical
    & Wide
    & $-0.660$ & $0.738$ \\
\quad Per-respondent ICL
    & Wide
    & $-2.939$ & $0.749$ \\
\quad Respondent-categorical
    & Wide with permutation augmentation
    & $-1.417$ & $0.432$ \\
\quad Per-respondent ICL
    & Wide with permutation augmentation
    & $-0.968$ & $0.745$ \\
\hline
\multicolumn{4}{l}{\emph{Pairwise TabPFN}} \\
\quad Respondent-categorical
    & Pairwise difference
    & $-0.807$ & $0.748$ \\
\quad Per-respondent ICL
    & Pairwise difference
    & $-0.797$ & $0.735$ \\
\quad Respondent-categorical
    & Pairwise ordered
    & $-0.811$ & $0.728$ \\
\quad Per-respondent ICL
    & Pairwise ordered
    & $-0.801$ & $0.735$ \\
\hline\hline
\end{tabular}
\end{table}

The heterogeneous results change the picture sharply. Moving from
homogeneous TabPFN to heterogeneity-aware TabPFN produces a much larger
improvement than moving across representation families within the
homogeneous setting. The best homogeneous TabPFN specification has
LL $=-1.115$, whereas the best heterogeneous specification, long
TabPFN with rank features and respondent-categorical encoding, reaches
LL $=-0.648$ and a hit rate of $0.772$. This improvement is also large
relative to the hierarchical benchmarks. The best HB specification, the
single-normal population model, achieves LL $=-0.725$ and hit rate
$0.745$; the more flexible finite-mixture and Dirichlet process
specifications do not improve on it in this dataset. Thus, the best
in-context TabPFN specification improves on the strongest HB benchmark
by approximately $8.0\%$ in predicted probability and $3.6\%$ in hit
rate. The best TabPFN configuration also runs in approximately $4.4$
seconds, against $70.4$
seconds for HB-MNL with the single-normal population, a sixteenfold
reduction in wall-clock time; the finite-mixture and Dirichlet process
specifications take comparable time to the single-normal specification.
This computational efficiency is itself a contribution of the
in-context approach: the cost of inference has been amortized into
pretraining, and no further MCMC is required at deployment.

These comparisons clarify what is, and is not, driving the gains.
The long, wide, and pairwise representations differ in performance, but
these differences are second-order relative to the effect of
introducing consumer heterogeneity. In the pooled setting, the best
TabPFN reformulation improves over pooled MNL by only about $3.9\%$ in
predicted probability. By contrast, adding consumer heterogeneity to the long-rank representation improves
predicted probability by approximately $60\%$ over the pooled long-rank baseline. This decomposition supports the interpretation developed in
Section~\ref{sec:choice-structure}: the primary challenge in adapting
off-the-shelf tabular prediction to panel choice data is not only the
set-valued nature of each choice occasion, but the cross-observation
dependence induced by persistent consumer preferences.

The two heterogeneity constructions should therefore be read primarily
as alternative ways of making consumer-level preference variation
available to the model. Their common pattern is more important than
their difference: once consumer heterogeneity is introduced,
predictive performance improves substantially across representation
families. Still, the two constructions are not identical.
Respondent-categorical encoding keeps all consumers in a shared
context, allowing the model to use population-level regularities while
learning consumer-specific differences through the categorical
feature. Per-respondent ICL isolates each consumer and therefore
relies more heavily on the amount of calibration history available for
that consumer. This distinction helps explain why per-respondent ICL
performs well for some long-format specifications but becomes unstable
for the wide representation, where the median consumer provides too
few context rows to support a multiclass prediction problem. The wide
representation is also the most sensitive to representation choice
within the heterogeneous setting: combining wide with the
respondent-categorical encoding and permutation augmentation
simultaneously produces results well below the pooled MNL baseline, and
we report these entries for completeness as evidence that some
refinements of the wide representation interact poorly with one another.

The aggregate comparison masks important variation across consumer
depth. Table~\ref{tab:icl_depth_bins} compares the strongest HB
benchmark against in-context TabPFN within each purchase-depth bin
under two specifications of TabPFN. The first holds the specification
fixed across all bins at the aggregate winner (long with rank
features and respondent-categorical encoding) so that a single
deployable model is evaluated everywhere. The second reports the best
TabPFN variant within each bin, where ``best'' is taken as the
variant with the highest holdout hit rate, with log-likelihood as a
tie-breaker among methods that reach the bin's hit-rate ceiling. The
two columns together let the reader distinguish what the fixed
aggregate winner gives up at a given depth from how much remains on
the table for variant-specific gains.

\begin{table}[h!]
\centering
\caption{In-context TabPFN versus HB by consumer purchase
depth.\protect\footnotemark}
\label{tab:icl_depth_bins}
\small
\begin{tabular}{l c rr rr rr}
\hline\hline
& & \multicolumn{2}{c}{\textbf{HB}}
  & \multicolumn{2}{c}{\textbf{TabPFN, fixed}}
  & \multicolumn{2}{c}{\textbf{TabPFN, best per bin}} \\
\cline{3-4}\cline{5-6}\cline{7-8}
\textbf{Bin} & \textbf{Occasions} & LL & Hit & LL & Hit & LL & Hit \\
\hline
B1 & 5--10  & $-1.010$ & $0.624$ & $-1.028$ & $0.570$ & $\mathbf{-1.003}$ & $\mathbf{0.634}$ \\
B2 & 11--20 & $-0.816$ & $0.687$ & $\mathbf{-0.625}$ & $\mathbf{0.798}$ & $\mathbf{-0.625}$ & $\mathbf{0.798}$ \\
B3 & 21--40 & $-0.440$ & $0.889$ & $\mathbf{-0.306}$ & $\mathbf{0.937}$ & $\mathbf{-0.306}$ & $\mathbf{0.937}$ \\
B4 & $>$40  & $\mathbf{-0.279}$ & $\mathbf{0.949}$ & $-0.351$ & $0.923$ & $-0.308$ & $\mathbf{0.949}$ \\
\hline
All & --- & $-0.725$ & $0.745$ & $-0.648$ & $0.772$ & --- & --- \\
\hline\hline
\end{tabular}
\end{table}
\footnotetext{The bin-mean log-likelihoods reported here are pulled by a
small set of consumers on which the two methods differ sharply. Five of
the 98 consumers contribute approximately $91\%$ of the aggregate
log-likelihood gap between the fixed TabPFN specification and HB (all
of them in B1--B3), and the median per-consumer log-likelihood gap is
small in every depth bin ($-0.05$ to $+0.06$) while the median hit-rate
gap is exactly zero in every bin. At the per-consumer level, HB wins
the log-likelihood comparison on 8 of the 13 consumers in B4. The best TabPFN variants per bin are: pairwise-difference with
antisymmetric-Borda aggregation under per-respondent ICL in B1; long
with rank features and respondent-categorical encoding in B2 and B3
(the same as the fixed specification); and long with rank features
under per-respondent ICL in B4. Several other heterogeneity-aware
TabPFN variants also reach the $0.9487$ hit-rate ceiling in B4,
matching HB on hit; among them, long with rank features under
per-respondent ICL has the highest log-likelihood. The aggregate row
is omitted for the best-per-bin column because no single TabPFN model
achieves all bin-specific bests simultaneously.}

Two broader patterns in the depth distribution are nonetheless clear. First, both HB and TabPFN benefit from deeper
consumer histories. HB improves monotonically across depth bins, from
LL $=-1.010$ in the shallow bin to LL $=-0.279$ in the deepest bin,
with hit rate rising from $0.624$ to $0.949$. TabPFN shows the same
broad pattern from shallow to medium-deep consumers: its LL improves
from $-1.028$ in B1 to $-0.625$ in B2 and $-0.306$ in B3, while hit
rate rises from $0.570$ to $0.798$ and then $0.937$. In the deepest
bin, however, TabPFN no longer improves, whereas HB continues to
benefit from the longer consumer histories. This suggests that the
two approaches use additional individual-level observations
differently: HB's parametric consumer-level posterior continues to
sharpen with depth, while the fixed-prior TabPFN specification appears
most useful in the middle of the depth distribution.

Second, TabPFN's aggregate advantage over HB is concentrated in the
medium-depth bins. In B2, TabPFN improves on HB by approximately
$21\%$ in predicted probability and $16\%$ in hit rate. In B3, TabPFN
improves on HB by approximately $14\%$ in predicted probability and
$5\%$ in hit rate. By contrast, HB outperforms TabPFN in the
shallowest bin and in the deepest bin on both metrics. Thus, the
overall TabPFN gain is not a uniform dominance result. Rather,
in-context TabPFN is most competitive when consumers have enough
history for the model to infer consumer-specific patterns, but not so
much history that the hierarchical Bayesian model can estimate
individual preferences very precisely.

This depth pattern is consistent with the unified shrinkage view. With very shallow
histories, the TFM's learned prior has limited consumer-specific information to condition
on, and HB's population pooling remains valuable. With medium-depth
histories, TabPFN can exploit the consumer identifier and rank-based
representation to extract useful individual-level signal, producing
its clearest advantage over HB. With deep histories, HB catches up or
surpasses TabPFN because the individual-level posterior is well
informed by each consumer's own data.

Overall, the in-context results show that the TFM's learned prior can be useful for
discrete choice prediction, but only
when the choice data are organized so that consumer heterogeneity is
visible to the model. Representation matters, especially the long
representation with simple within-set rank features, but the main
empirical result is the large jump from homogeneous to
heterogeneity-aware prediction. The depth-bin analysis further shows
that this advantage is concentrated in the middle of the
consumer-depth distribution. This pattern motivates the fine-tuning
analysis in the next subsection: if the shallow bin is precisely where
the learned prior has too little consumer-specific information to condition on, then
updating the prior using population choice data should be most valuable there.

\subsection{TabPFN Fine-Tuning Results}
\label{subsec:results-ft}

The preceding subsection evaluates TabPFN with fixed pretrained weights.
We now ask whether performance improves when the learned prior is updated using the focal
choice population. Conceptually,
fine-tuning differs from ICL because population
information enters the model weights before prediction. This makes
fine-tuning the closest TabPFN analogue to the population-learning step
in HB, where information from all consumers is used to estimate the
population distribution that anchors individual-level predictions.

A practical constraint is that classifier fine-tuning is currently
supported for TabPFN v2.5, whereas the strongest in-context results in
the previous subsection use TabPFN v2.6. We therefore report three
quantities: v2.5 ICL, v2.5 fine-tuning, and v2.6
ICL. The comparison between v2.5 ICL and v2.5
fine-tuning isolates the effect of fine-tuning holding model version
fixed. The comparison with v2.6 ICL shows how much of the performance
gap to the newer off-the-shelf model is closed by fine-tuning.
Table~\ref{tab:ft_aggregate} reports these comparisons for four
representative specifications.

\begin{table}[h!]
\centering
\caption{Aggregate fine-tuning results.}
\label{tab:ft_aggregate}
\small
\begin{tabular}{l rr rr rr}
\hline\hline
& \multicolumn{2}{c}{\textbf{v2.5 ICL}}
& \multicolumn{2}{c}{\textbf{v2.5 FT}}
& \multicolumn{2}{c}{\textbf{v2.6 ICL}} \\
\cline{2-3}\cline{4-5}\cline{6-7}
\textbf{Method} & LL & Hit & LL & Hit & LL & Hit \\
\hline
Long, pooled
& $-1.124$ & $0.469$
& $\mathbf{-1.105}$ & $0.442$
& $-1.115$ & $\mathbf{0.483}$ \\

Long, respondent-categorical
& $-0.686$ & $0.752$
& $-0.679$ & $\mathbf{0.762}$
& $\mathbf{-0.672}$ & $\mathbf{0.762}$ \\

Long, per-respondent ICL
& $-0.721$ & $0.755$
& $\mathbf{-0.716}$ & $0.735$
& $-0.725$ & $\mathbf{0.765}$ \\

Long-rank, respondent-categorical
& $-0.720$ & $0.728$
& $-0.683$ & $0.769$
& $\mathbf{-0.648}$ & $\mathbf{0.772}$ \\
\hline\hline
\end{tabular}
\end{table}

The aggregate results show that fine-tuning helps, but does not
uniformly dominate the newer off-the-shelf model. Holding the model
version fixed at v2.5, fine-tuning improves log-likelihood for all
four specifications, with gains in predicted probability ranging from
$0.5\%$ to $3.8\%$. The largest gain occurs for the long-rank
respondent-categorical specification, where fine-tuning improves
predicted probability by $3.8\%$ and hit rate by $5.6\%$. However,
v2.6 ICL remains the strongest aggregate specification for the main
winner, with LL $=-0.648$ and hit rate $=0.772$. For this method, the
v2.6 upgrade improves predicted probability by $7.5\%$ relative to
v2.5 ICL, while fine-tuning v2.5 improves it by $3.8\%$, closing
roughly half of the version gap.

The pattern is consistent with the prior-updating interpretation.
Fine-tuning is most useful when the model has a channel through which
population-level choice regularities can be expressed. The
respondent-categorical specifications allow the model to learn how
consumer identity relates to choice behavior within a shared
population context. By contrast, fine-tuning provides only modest
calibration gains for the pooled and per-respondent specifications, and
its effects on hit rate are mixed. This suggests that fine-tuning is
not simply adding generic flexibility; it is most valuable when the
representation gives the updated prior a way to organize
population-level heterogeneity.

The aggregate comparison, however, does not fully reveal where
fine-tuning is useful. The depth-bin analysis in the previous
subsection suggests that the shallow bin is the most prior-dependent
regime: consumers in B1 have too little calibration history for
consumer-specific patterns to be reliably inferred from context alone.
If fine-tuning moves TabPFN toward a population choice-aware prior, its
effect should therefore be largest in B1.
Table~\ref{tab:ft_bins_lrrc} tests this implication for the strongest
overall TabPFN specification, long with rank features and
respondent-categorical encoding.

\begin{table}[h!]
\centering
\caption{Fine-tuning results by consumer depth for long-rank
respondent-categorical TabPFN.}
\label{tab:ft_bins_lrrc}
\small
\begin{tabular}{l rr rr rr rr}
\hline\hline
& \multicolumn{2}{c}{\textbf{HB}}
& \multicolumn{2}{c}{\textbf{v2.5 ICL}}
& \multicolumn{2}{c}{\textbf{v2.5 FT}}
& \multicolumn{2}{c}{\textbf{v2.6 ICL}} \\
\cline{2-3}\cline{4-5}\cline{6-7}\cline{8-9}
\textbf{Bin} & LL & Hit & LL & Hit & LL & Hit & LL & Hit \\
\hline
B1
& $-1.010$ & $0.624$
& $-1.033$ & $0.527$
& $\mathbf{-0.993}$ & $\mathbf{0.645}$
& $-1.028$ & $0.570$ \\

B2
& $-0.816$ & $0.687$
& $-0.822$ & $0.707$
& $-0.778$ & $0.727$
& $\mathbf{-0.625}$ & $\mathbf{0.798}$ \\

B3
& $-0.440$ & $0.889$
& $-0.317$ & $0.921$
& $\mathbf{-0.297}$ & $0.905$
& $-0.306$ & $\mathbf{0.937}$ \\

B4
& $\mathbf{-0.279}$ & $\mathbf{0.949}$
& $-0.370$ & $0.949$
& $-0.332$ & $0.949$
& $-0.351$ & $0.923$ \\

\hline
All
& $-0.725$ & $0.745$
& $-0.720$ & $0.728$
& $-0.683$ & $0.769$
& $\mathbf{-0.648}$ & $\mathbf{0.772}$ \\
\hline\hline
\end{tabular}
\end{table}

The depth-bin results sharpen the interpretation. In the shallow bin,
fine-tuned v2.5 is the best model on both metrics, beating HB, v2.5
ICL, and v2.6 ICL. This is exactly the regime in which consumer-level
context is thinnest and predictions rely most heavily on the prior.
Fine-tuning improves B1 hit rate by approximately $22\%$ relative to
v2.5 ICL and by $13\%$ relative to v2.6 ICL. It also improves B1
predicted probability relative to both fixed-prior TabPFN variants and
HB. The B1 advantage is attributable to fine-tuning rather than to a
confound with the newer off-the-shelf model: in this bin, the version
upgrade from v2.5 ICL to v2.6 ICL contributes only about $0.5\%$ in
predicted probability, whereas fine-tuning within v2.5 contributes
about $4.1\%$. This pattern suggests that updating the pretrained
prior with population choice data is especially valuable when
consumer-specific histories are too sparse for ICL
alone.

The advantage of fine-tuning is less pronounced in deeper bins. In B2,
v2.6 ICL dominates both HB and fine-tuned v2.5. In B3, all TabPFN
variants perform well relative to HB, with fine-tuned v2.5 slightly
ahead on LL and v2.6 ICL ahead on hit rate. In B4, HB regains the
advantage in LL and ties or leads on hit rate, consistent with the
depth analysis above: when each consumer has a long history, the
hierarchical model can estimate individual preferences precisely. Thus,
fine-tuning does not uniformly dominate either HB or off-the-shelf
TabPFN. Its contribution is concentrated where the theory predicts it
should be: among consumers with sparse histories, where
population-level regularization is most valuable.

Taken together, the fine-tuning results qualify and extend the
in-context findings. Off-the-shelf TabPFN v2.6 remains the best
aggregate model in this application, but fine-tuning v2.5 substantially
improves over v2.5 ICL and is especially effective in the shallowest
consumer bin. This supports the interpretation of fine-tuning as
moving TabPFN from a learned prior toward a population choice-aware prior. ICL conditions on the observed
calibration choices; fine-tuning changes the prior that the model
brings to those choices. The two operations therefore play different
roles in adapting TFMs to panel choice data.

\section{Conclusion}
\label{sec:conclusion}

This paper studies the application of TFMs to discrete choice problems and identifies a fundamental structural mismatch between the two. Discrete choice is inherently set-valued and heterogeneous, with outcomes defined by comparisons within choice sets and linked across observations through individual-level preferences. In contrast, TFMs are built on a row-wise learning paradigm that imposes no explicit
relational or set-valued structure. This mismatch is not a matter of model capacity, but of representation: the economic structure of choice is not directly aligned with the inductive bias of tabular learning. As a result, direct applications of TFMs yield performance comparable to pooled models that do not capture heterogeneity or within-set competition.

To address this, we propose a structured reformulation of discrete choice prediction that explicitly aligns data representation and task construction with the underlying decision structure. By encoding both choice-set dependence and individual-level heterogeneity within a row-based framework, the reformulation enables pretrained TFMs to approximate the original set-conditioned prediction problem. Under this alignment, TFMs match or exceed the predictive performance of established heterogeneity-aware DCMs, with the strongest aggregate results achieved via ICL and additional gains available through fine-tuning on population choice data, particularly for consumers with sparse purchase histories.

Heterogeneity encoding is the dominant driver of predictive improvement,
with choice-set representation providing a secondary but consistent contribution. By conditioning predictions on respondent identity (either
through an explicit indicator feature or through per-respondent in-context
learning), TFMs recover personalized patterns without explicit parameter estimation or
MCMC. The appropriate form of this encoding depends on data depth: respondent-indicator
approaches are more effective when per-consumer histories are moderately rich, while
fine-tuning on population choice data is most
effective in sparse settings, providing the largest gains where
consumer-specific histories are too thin for ICL to reliably infer
individual preferences.
Empirically, TFMs offer the greatest aggregate advantage
over HB in the medium-data regime (roughly 10--40 observations per consumer), where HB’s
parametric normal prior is most likely to bind for atypical consumers; the two methods
converge in both sparse and data-rich settings.
Managerially, our findings suggest that TFMs are an effective approach for individual-level preference prediction in the sparse-data settings that dominate practice, where each consumer supplies only a handful of purchase occasions.

The unified shrinkage view developed in Section~\ref{sec:tfm-shrinkage} provides the theoretical grounding for this alignment: TFMs and HB estimation occupy adjacent positions in a common methodological progression from pooled to individual-level inference, differing primarily in how the prior over consumer preferences is acquired (learned from synthetic data during pretraining) and updated (conditioned in-context or via fine-tuning, rather than via MCMC). Rather than direct replacements for existing approaches, TFMs are better understood as prediction-oriented tools whose performance depends on how the problem is formulated, offering a computationally lightweight alternative that avoids task-specific estimation while retaining competitive accuracy. More broadly, where outcomes are defined over sets and linked across observations, structure cannot be left for the model to infer implicitly; it must be encoded in the representation. Discrete choice provides a clear example, but the same tension is likely to arise in other decision-making settings with relational and hierarchical structure.

Several directions emerge from this work. Within the choice modeling context,
the dominant role of heterogeneity encoding identifies richer preference representation
as the highest-leverage near-term extension: latent class and mixed-membership specifications
within an ICL framework would capture the discrete segmentation structure
that many marketing applications require.
Looking beyond prediction, the most important extension is toward
counterfactual demand analysis: price elasticities, market share responses to new
products, and welfare implications of assortment decisions. Adapting PFN architectures to
causal settings, pretraining on causal data-generating processes to enable zero-shot
estimation of treatment effects without specifying a parametric demand system, is a
natural next step. The rapid development of the TFM literature itself opens a further
agenda. Recent architectures scale ICL to hundreds of thousands of observations
\citep{priorlabs2025tabpfn25,qu2025tabicl}, making TFMs applicable to the large
transaction logs that characterize commercial demand estimation and revenue management,
and multi-table relational extensions \citep{fey2024relational,eremeev2025turning} would
both leverage the database structures firms already maintain and reduce reliance on the
external reformulation the present work provides. Domain-specific pretraining is a further
frontier: models pretrained on curated real-world
tables have shown substantial improvements over models trained only on synthetic priors \citep{ma2025tabdpt},
and pretraining on a retailer's transaction history could yield a demand-specialized TFM
that transfers across product categories. Finally,
extracting price elasticities and attribute importances from TFM predictions, without
imposing parametric structure on the demand function, would make the approach actionable
for the pricing, assortment, and targeting decisions that motivate choice modeling in the
first place.

\bibliographystyle{ormsv080}
\bibliography{tfm}

\newpage 
\appendix

\section{The Hierarchical Bayesian Framework for Discrete Choice}
\label{app:hb}

This appendix provides a self-contained account of the hierarchical Bayesian (HB)
framework for discrete choice estimation. HB is the dominant method in marketing practice
for recovering individual-level preference parameters from panel choice data, and it serves
as the primary benchmark in the empirical analysis of the main text.

\subsection*{The Challenge: Limited Individual-Level Data}

Estimating individual-level consumer preferences is a central challenge in marketing
research. In conjoint studies, a typical respondent completes 10 to 20 choice tasks; in
scanner panel data, a household may have a few dozen purchase occasions. Either way, the
number of observations per person is far too small to estimate individual preference
parameters reliably without borrowing strength from other consumers. A model fit to one
person's data alone would be highly unstable, overfitting idiosyncratic noise rather than
recovering genuine preferences.

\subsection*{The Two-Level Model}

The HB approach resolves this by modeling individual parameters as draws from a common
population distribution \citep{rossi2005bayesian,allenby1998marketing}. The model has two
levels.

At the \emph{individual level}, consumer $i$ makes choices according to a utility
maximization rule. On choice occasion $t$, alternative $j$ is chosen from set
$S_t$ if
\begin{equation}
    u_{ijt} = \mathbf{x}_{ijt}'\boldsymbol{\beta}_i + \varepsilon_{ijt}
    > u_{ikt} \quad \text{for all } k \neq j \text{ in } S_t,
    \label{eq:app_utility}
\end{equation}
where $\mathbf{x}_{ijt}$ is a vector of observed alternative attributes,
$\boldsymbol{\beta}_i$ is an individual-specific coefficient vector (part-worths), and
$\varepsilon_{ijt}$ is an idiosyncratic error term. Assuming i.i.d.\ Type~I extreme value
errors, the probability that consumer $i$ chooses alternative $j$ from $S_t$ takes the
multinomial logit form:
\begin{equation}
    P(j \mid S_t;\, \boldsymbol{\beta}_i)
    = \frac{\exp(\mathbf{x}_{ijt}'\boldsymbol{\beta}_i)}
           {\sum_{k \in S_t} \exp(\mathbf{x}_{ikt}'\boldsymbol{\beta}_i)}.
    \label{eq:app_mnl}
\end{equation}

At the \emph{population level}, individual parameters are assumed to follow a multivariate
normal distribution,
\begin{equation}
    \boldsymbol{\beta}_i \sim \mathcal{N}(\boldsymbol{\mu}, \boldsymbol{\Sigma}),
    \label{eq:app_hb_prior}
\end{equation}
with population mean $\boldsymbol{\mu}$ and covariance $\boldsymbol{\Sigma}$ estimated
jointly with all individual parameters from the full panel. The posterior distribution
over $\boldsymbol{\beta}_i$ is obtained by combining the individual choice likelihood
implied by \eqref{eq:app_mnl} with the population prior in \eqref{eq:app_hb_prior}.

\subsection*{Shrinkage and Estimation}

The key property of the posterior mean estimate is \emph{shrinkage}: the estimate for
consumer $i$ can be written approximately as a weighted average of the individual's own
maximum likelihood estimate and the population mean $\boldsymbol{\mu}$, with weights
determined by how many observations that person contributed. A consumer with few
observations receives a posterior pulled strongly toward $\boldsymbol{\mu}$; a consumer
with many observations receives one close to their own choice history. Information flows
in both directions: each individual's choices update the population distribution, and the
population distribution regularizes each individual's estimate. This adaptive pooling is
what allows HB to produce stable individual-level estimates even from short purchase
histories.

Estimation proceeds via Markov chain Monte Carlo (MCMC), which yields full posterior
distributions and individual-level parameter draws. These draws support downstream tasks
such as consumer segmentation, targeting, and assortment optimization. HB has been widely
adopted in academic marketing research and commercial software
\citep{rossi1996value,train2009discrete}, and represents the state of practice for conjoint
and scanner panel applications.

\subsection*{Relationship to Tabular Foundation Models}

The TFM reformulation developed in the main text pursues the same goal as HB: recovering
reliable individual-level choice predictions from limited per-consumer data. The
mechanisms differ. In HB, pooling is explicit and parametric: the researcher specifies the
prior, MCMC recovers the full posterior, and the amount of shrinkage each consumer
receives is transparent. In a TFM, pooling is implicit and amortized: a prior over
data-generating processes is learned during pretraining on synthetic data, and
individual-level adaptation occurs through the examples supplied in context at prediction
time, without any gradient updates. The cost is computational speed (a single forward pass
replaces MCMC); the cost is that the learned prior is not directly interpretable.

\end{document}